\documentclass[letterpaper, 10 pt, conference]{ieeeconf}  %

\IEEEoverridecommandlockouts                              %

\title{\LARGE \bf
Dynamics Models in the Aggressive Off-Road Driving Regime
}

\author{Tyler Han, Sidharth Talia, Rohan Panicker, Preet Shah, Neel Jawale, Byron Boots \\
University of Washington
\thanks{This material is based upon work supported by the National Science Foundation Graduate Research Fellowship under Grant No. DGE-2140004. Any opinions, findings, and conclusions or recommendations expressed in this material are those of the author(s) and do not necessarily reflect the views of the National Science Foundation.}
\thanks{BeamNG simulator interface and model code can be found at: (\url{https://github.com/prl-mushr/BeamNGRL}). Dataset can be found at: \url{https://tinyurl.com/5n9y4akc}}%
}

\usepackage{graphicx} %
\usepackage{amsmath} %
\usepackage{amssymb}  %
\usepackage{booktabs}
\usepackage{mathtools}
\usepackage[style=ieee, url=false, doi=false]{biblatex}
\usepackage{hyperref}
\usepackage{cleveref}
\usepackage{multirow}
\usepackage[table]{xcolor}
\usepackage{siunitx}

\DeclareMathOperator*{\argmin}{arg\,min}

\DeclareMathOperator*{\hmne}{{MaxNormedError}_H}

\begin{document}

\renewcommand{\vec}[1]{\ensuremath{\mathbf{#1}}}
\newcommand{\xb}{\vec x}
\newcommand{\phib}{\boldsymbol{\phi}}
\newcommand{\xdmax}{\dot{\xb}_\text{max}}
\newcommand{\xdmin}{\dot{\xb}_\text{min}}
\newcommand{\norm}[1]{\left\lVert#1\right\rVert}
\newcommand{\abs}[1]{\left\lvert#1\right\rvert}
\newcommand{\Dagg}{\mathcal{D}_\text{agg}}
\newcommand{\std}[1]{\tiny{$\pm$ #1}}

\maketitle
\thispagestyle{empty}
\pagestyle{empty}

\begin{abstract}

Current developments in autonomous off-road driving are steadily increasing performance through higher speeds and more challenging, unstructured environments. However, this operating regime subjects the vehicle to larger inertial effects, where consideration of higher-order states is necessary to avoid failures such as rollovers or excessive impact forces.
Aggressive driving through Model Predictive Control (MPC) in these conditions requires dynamics models that accurately predict safety-critical information.
This work aims to empirically quantify this aggressive operating regime and its effects on the performance of current models. We evaluate three dynamics models of varying complexity on two distinct off-road driving datasets: one simulated and the other real-world. By conditioning trajectory data on higher-order states, we show that model accuracy degrades with aggressiveness and simpler models degrade faster. These models are also validated across datasets, where accuracies over safety-critical states are reported and provide benchmarks for future work. 
\end{abstract}
\section{INTRODUCTION}

In off-road autonomous driving, Model Predictive Control (MPC) is widely used for fast planning and control to address problems such as obstacle avoidance or uneven terrain. Historically, MPC has performed well in practice, accommodating unknown and approximate dynamics through frequent model prediction and re-optimization \cite{han_model_2023, williams_information_2017, gibson_multi-step_2023, pagot_real-time_2020}.
However, the effectiveness of this approach critically depends on the accuracy of the model prediction step. Particularly in unstructured environments, MPC methods often suffer due to models that inadequately capture safety-critical dynamics \cite{williams_information_2017, kabzan_learning-based_2019}.
Nonetheless, aggressive driving aims to maximize system performance without inducing critical failure modes.
In unstructured off-road driving, uneven and nonplanar terrain geometries can cause destabilizing impact forces, wheel airtime, and loss of control \cite{lee_learning_2023, han_model_2023}. While driving slower can prevent these failures, an understanding of vehicle dynamics can mitigate them even at high speeds \cite{han_model_2023}.  
\begin{figure}[t]
    \centering
    \includegraphics[width=.85\linewidth]{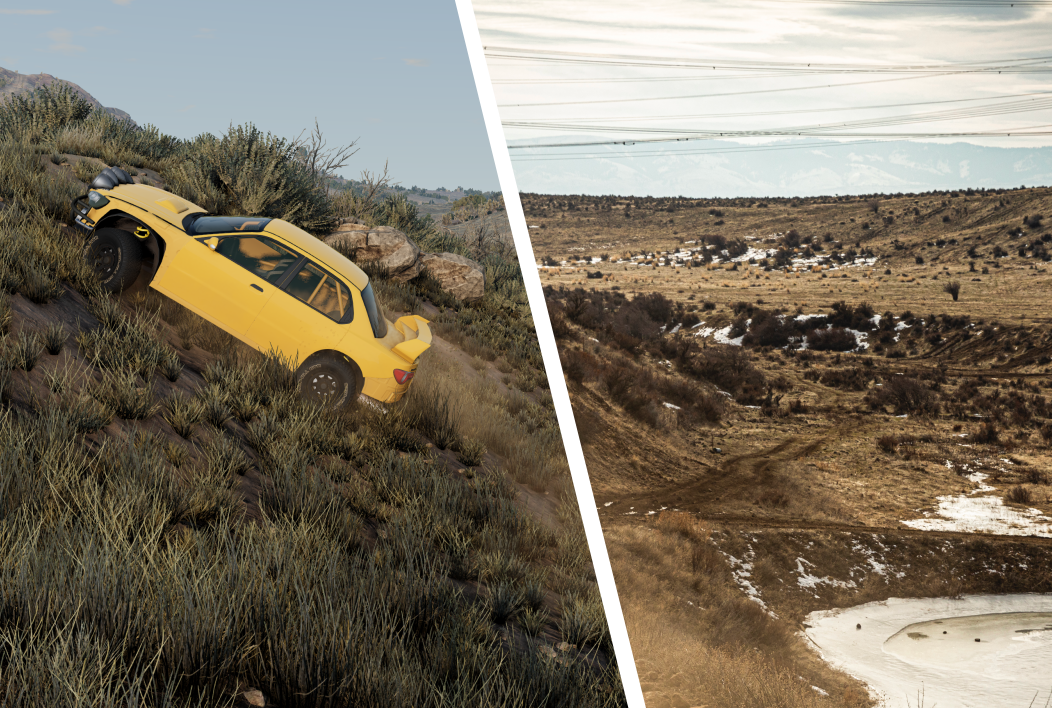}
    \caption{This work evaluates dynamics models of varying complexity in challenging and unstructured environments, both simulated (BeamNG, left) and real (Washington, right).}
    \vspace{-15pt}
    \label{fig:ellensburg}
\end{figure}
Prior works have not extensively studied this regime for off-road driving.

The NATO Reference Mobility Model (NRMM) has been historically used to characterize ground traversability and difficulty \cite{carruth_challenges_2022, nrmm, goodin_fast_2021}. 
These models perform static two-dimensional analyses over distinct ground shapes and provide estimates of vertical accelerations and their effect on expected traversal speeds. 
However, in the aggressive driving regime, speeds change quickly, ground geometry varies as a continuum, and inertial forces are exerted in all three dimensions.

In addition, factors such as complex steering mechanisms, suspension, tire slippage, diverse driving conditions, and deformable terrain can make analytical dynamics models an extensive
engineering effort.
Recently, researchers have employed learned dynamics models to address these issues \cite{gibson_multi-step_2023, maheshwari_piaug_2023, kim_physics_2022}. In these works, various model architectures are validated against either positional or velocity errors in both simulated and real environments. As the field advances towards more aggressive driving and maneuvers, we stress that validation on only these states is not sufficient. 

Off-road driving, or traversability in general, is a complex problem that fundamentally depends upon both perception and control. For this work, we limit our scope, assuming reliable perception, to better study aggressiveness for dynamics modeling.
In the real world, we use \cite{terrainnet} to achieve this, while in simulation the perception is the ground truth obtained directly from the simulator.

We demonstrate that off-road driving is an increasingly difficult dynamics modeling problem in the aggressive regime.
Moreover, we provide results using analytic and learned dynamics models on off-road driving datasets. Our contributions are:
\begin{enumerate}
    \item \textbf{A quantitative study of Aggressiveness} and its effect on analytic (no-slip and slip) and learned dynamics models for off-road driving.
    \item \textbf{Safety-critical evaluation of dynamics models} on simulated (BeamNG) and real (Washington) aggressive driving data with steep slopes and banks, high speeds, vegetation, and diverse terrain.
\end{enumerate}

\section{Literature Review}

Aggressive control behavior has been previously explored for on-road autonomous vehicles, small-scale off-road vehicles, drones, and manipulators \cite{andersson_aggressive_1989, arab_motion_2023, mellinger_trajectory_2012, williams_aggressive_2016}. Interestingly, both quantitative and qualitative definitions vary from one work to the next. For instance, Mellinger et al. \cite{mellinger_trajectory_2012} suggest trajectories that are momentarily unstable (due to some agile maneuver) but recover to a stable state are aggressive. On the other hand, \cite{arab_motion_2023} provides a rigorous definition for a small-scale wheeled robot, proposing a safety region defined by the stability limits of an LQR steering controller. All works generally do agree on some fundamental characteristics of aggressive control behavior: the autonomy operates at high speeds, maintains safety, and can saturate actuators reliably. We note that in the off-road setting, stability can be impractical to model due to highly nonlinear dynamics and diverse conditions \cite{williams_aggressive_2016, knaup_safe_2023, terrainnet, han_model_2023}. 

Rather than studying stability conditions, failure modes can instead be modeled.
For instance, there is substantial work in rollover prevention in the automotive safety testing industry.
Lateral force modeling is used to predict and prevent rollover failures through metrics such as the static stability factor, lateral load transfer ratio, and time to rollover \cite{chen_differential-braking-based_2001, doumiati_lateral_2009, huston_another_2014}. As proposed in \cite{han_model_2023, lee_learning_2023}, modeling additional components of acceleration and angular rates is critical to failure prevention in the aggressive off-road regime. By more rigorously defining this regime in this work, we study its impact on various dynamics model classes.

Learning dynamics for robot modeling and optimal control has also been extensively studied \cite{nguyen-tuong_model_2011}. To make model learning more tractable and efficient, prior work often takes advantage of known holonomic constraints through hybrid models that combine kinematic and learned functions.
This approach has been shown to demonstrate good experimental results including complex applications like drifting for a small-scale off-road platform \cite{williams_information_2017}.
Gibson et al. employ an LSTM to learn a multi-step predictive model to be used in optimal real-time sample-based control for multiple environments \cite{gibson_multi-step_2023}. However, the model is validated using only positional and yaw accuracy.  
Maheshwari et al. \cite{maheshwari_piaug_2023} learn a model that is trained on low-velocity data and demonstrates generalization to higher-velocity test scenarios. The authors also demonstrate the importance of predicting slippage, showing a large divergence in aggressive flat turns between a kinematic and learned model.

\section{Dynamics models}
For car-like vehicles with Ackermann steering mechanisms \cite{ackermann}, there exist many models, such as the point mass model, bicycle model, as well as the four-wheel model \cite{dynamic_gerdes}. 
Of these, we consider the bicycle model class.
Within this class, we consider the No-slip model, slip-based models, and Terrain-conditioned models.
These models at their core describe body-frame rates, such as body frame velocity, rotation rates, or accelerations. These accelerations are then transformed into the world frame using the vehicle orientation, and gravity and other inertial effects are added using the predicted or derived rotation rates. 
For brevity, we refer the reader to the work by \cite{non_planar_anal} which describes these transformations and compensations in detail.
\begin{figure}[!t]
\centering
\includegraphics[width=0.9\linewidth]{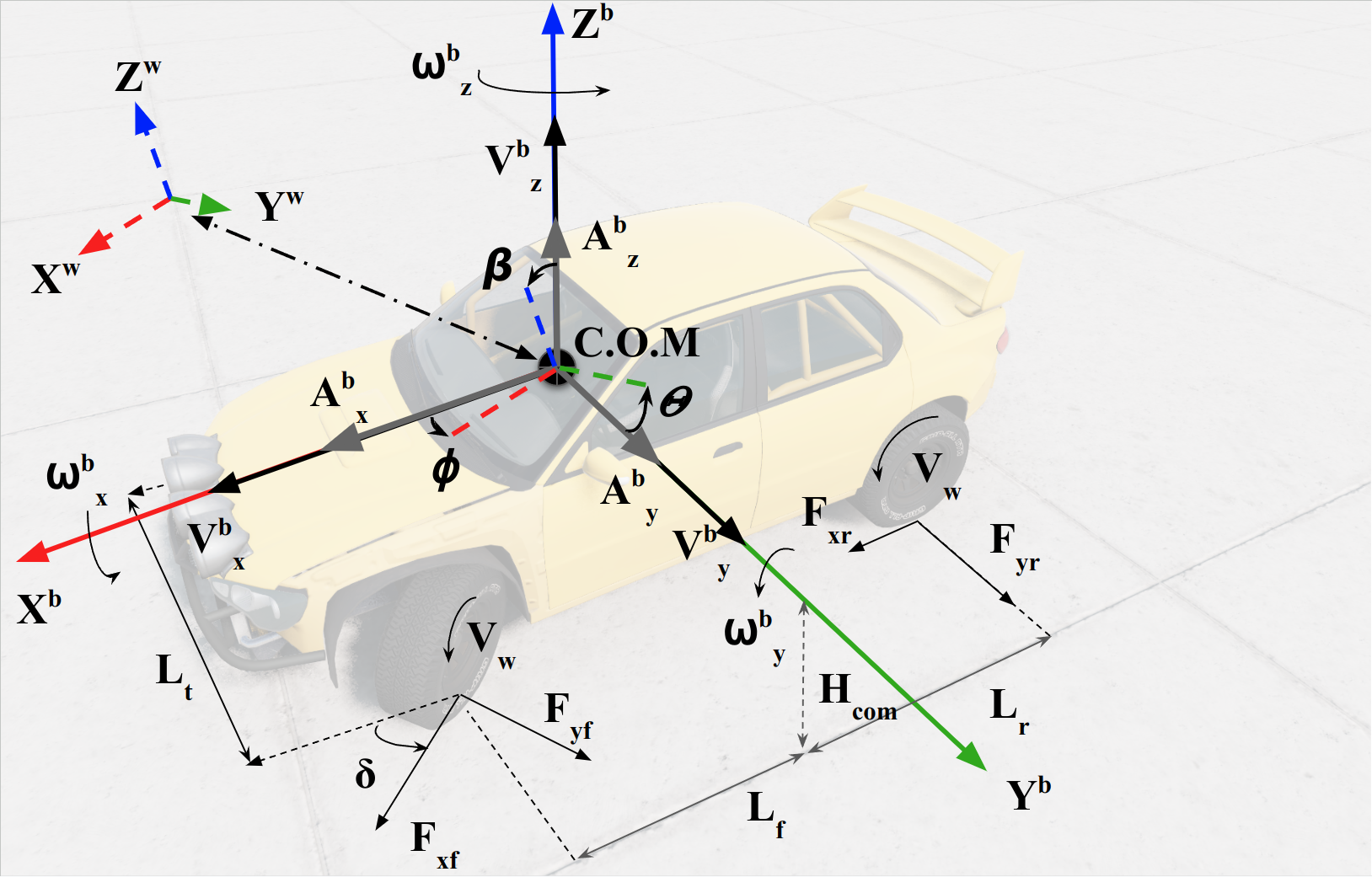}
\caption{Illustration of the coordinate frame used by dynamics models}
\label{fig:coordinate_frame}
\vspace{-15pt}
\end{figure}

\subsubsection{SE(3) no-slip based bicycle model}
This model refers to the no-slip approximate bicycle model 
with its velocities and positions projected in 3D using the elevation map. Prior works such as \cite{han_model_2023, maheshwari_piaug_2023} have also used such a model.
Here:
\begin{align}\label{noslip_eq}
\begin{split}
    V^b_x &= V_w, V^b_y = 0, V^b_z = 0 , \quad \omega_z = V^b_x * \tan(\delta)/L_{fr}
\end{split}
\end{align}
Where $V_w$ represents the wheel speed, $V^b$ represents the body frame velocity, $\omega_z$ represents the body frame rotation rate around the Z axis, $\delta$ represents the steering angle in radians, and $L_{fr}$ represents the wheelbase of the car.
\subsubsection{SE(3) slip based approximate model}
The slip-based approximate model uses a single-track bicycle model as described by \cite{alternate_tire_model_evidence}, which considers non-planar surfaces, with a simplified Pacejka tire model~\cite{simplified_pacejka}.

The body frame Forces $F^b_x, F^b_y, F^b_z$ and rotational acceleration $\dot{\omega}^b_z$ are given by:
\begin{align}\label{wheel_vel}
\begin{split}
F^b_x &= F_{xr} + F_{xf} \cos(\delta) - F_{yf} \sin(\delta) + m g \sin(\theta) \\
F^b_y &= F_{yr} + F_{yf} \cos(\delta) + F_{xf} \sin(\delta) + m g \sin(\phi) \\
F^b_z &= m (g \cos{\beta} - V_x \omega_y + V_y \omega_x)\\
\dot{\omega}^b_z &= ((F_{xf} \sin(\delta) + F_{yf} \cos(\delta))L_f - F_{yr} L_r)/J_z
\end{split}
\end{align}
Where the terms denoted by $F$ and $L$ represent forces and lengths respectively, described in Fig\ref{fig:coordinate_frame}, $m$ represents mass, and $J_z$ represents the mass moment of inertia around its Z axis.
These lateral and longitudinal accelerations are then used to update the body frame velocity $V^b$, which is transformed into the world frame to update the position. We omit details regarding exact implementation for brevity and refer the reader to prior works such as \cite{hound, non_planar_anal} that have also used such a model.

\subsubsection{Terrain conditioned dynamics modeling}
We consider the approach presented in the work by \cite{terrainCNN} as our fully-learned baseline. 
However, for simplicity, we do not re-implement a probabilistic ensemble (PE) dynamics model \cite{probabilistic_dynamics} and do not consider the history of the past states. 
We believe this version represents the core idea and note that it performs better than the analytical baselines.
In this approach, the model takes the patch of terrain directly under the vehicle $P$, the vehicle's intrinsic state and control inputs, and learns to predict the change in velocity and rotation rates for the next timestep by taking a single step loss on the velocity and rotation rates.
\begin{align}\label{wheel_vel}
\begin{split}
\Delta V^b, \Delta \omega^b = f(V^b, \omega^b, \phi,\theta, P, \delta, V_w)
\end{split}
\end{align}
During execution, the predicted delta velocity and rotation rates are integrated to obtain the velocity, orientation, and position using the ``Explicit Kinematic Layer'' in \cite{terrainCNN}.

\section{Aggressiveness}

\subsection{Definition} \label{sec:aggressiveness}

As prior works have corroborated qualitatively, aggressiveness is generally considered to be operating at the dynamic limits of the vehicle \cite{williams_aggressive_2016, andersson_aggressive_1989, mellinger_trajectory_2012, knaup_safe_2023}.
Following this intuition, we define aggressive driving to be trajectories constrained by these dynamic limits, or higher-order states (e.g. velocity, acceleration, etc.). Let $\xdmin$ and $\xdmax$ be lower and upper bounds on the differentiated state $\dot{\xb}_{t}$ at time $t$. 
\begin{align}\small
    \tau \coloneqq& \begin{bmatrix}
        \vec x_{t_1} & \vec x_{t_1} & \cdots & \vec x_{t_H} 
    \end{bmatrix} \\
    \text{s.t.} \qquad & \dot{\xb}_\text{min} \preceq \dot{\xb}_{t_i} \preceq \dot{\xb}_\text{max}, \quad i=1,\dots,H \label{dbox} \\
    &\tau = \argmin_{\tau' \in \Phi} C(\tau') \label{eq:optimality}
\end{align}
where the cost function $C(\cdot)$ in \eqref{eq:optimality} represents a combination of competing objectives, such as moving toward a goal and/or avoiding obstacles. In practice, this competing objective will make \eqref{dbox} tight \cite{han_model_2023, lee_learning_2023, fan_step_2021, williams_aggressive_2016}.
Without assuming knowledge about $C(\cdot)$, our primary interest is in box constraint \eqref{dbox}. These constraints represent fundamental constraints on the system, be it mechanical or human tolerable limits.
For off-road driving, constraints must be placed on several higher-order states to reliably maintain control and safety in the aggressive regime. For example, lateral accelerations are limited by the critical rollover acceleration \cite{doumiati_lateral_2009, chen_differential-braking-based_2001, huston_another_2014} and angular velocities and vertical accelerations are limited by airtime and impact tolerances \cite{han_model_2023, lee_learning_2023}.

\subsection{Measuring Aggressiveness with Free Energy} \label{sec:energy}

Without inductive biases, expressive model classes generally require a diverse dataset that provides adequate coverage of the relevant state space. However, there are a few complications with seeking coverage of \eqref{dbox} outright; including information about the geometry and semantics can result in exploding dimensionality, parts of the constrained space may not be reachable or observable at all, or the bounds $\xdmin$ and $\xdmax$ may not be known or can be subject to the data collection policy's preferences.
To evaluate whether a trajectory $\tau$ improves coverage of our constrained state space, we employ the free-energy function \cite{liu_energy-based_2020} to score $\tau$ using its distance from $\mathcal{D}$ as determined by distributions $p_i$,
\begin{align}
    E(\tau) = -T \cdot \log{ \sum^K_i \exp{ \left\{ (p_i \circ \phib[i])(\tau) / T \right\} } } \label{aggressiveness}
\end{align}
where $T$ is the temperature hyperparameter and $\phib : \mathbb{R}^{n \times H} \rightarrow \mathbb{R}^K$ is a feature function that captures the observed bounds,
\begin{small}
    $\phib(\tau) = \begin{bmatrix}
    \max_t{\dot{\xb}_t[k_1]} & \min_t{\dot{\xb}_t[k_1]} & \hdots &
    \max_t{\dot{\xb}_t[k_d]} & \min_t{\dot{\xb}_t[k_d]}
\end{bmatrix}$.
\end{small}
Note that we choose our state components through $k_i$ rather than include all states to reduce feature space dimensionality and noise.

\section{Experimentation}

\begin{figure}[t]
    \centering
    \includegraphics[width=.9\linewidth]{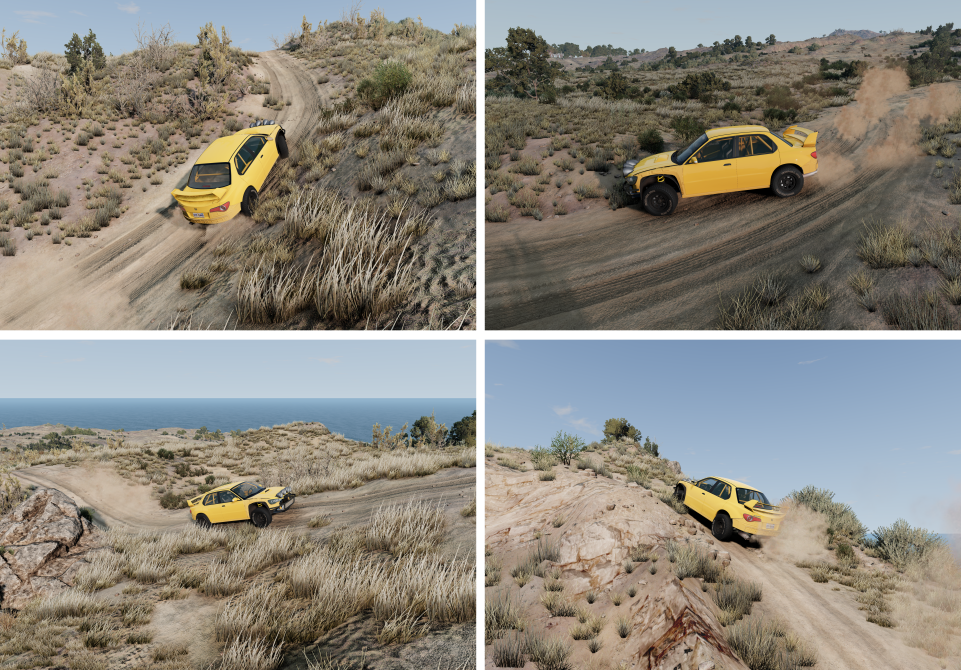}
    \caption{Freeze-frames of aggressive trajectories in BeamNGRL dataset.}
    \label{fig:beamng}
    \vspace{-10pt}
\end{figure}

\subsection{Aggressiveness and Dynamics Models} \label{sec:result1}

\begin{table*}[t]
    \centering
    \caption{Model Performance on Datasets Across States}
    \begin{tabular}{llrrrrrrr}
    \toprule
       & & \multicolumn{7}{c}{$H$-Max Normed Error ($H$-MNE)} \\
       \cmidrule(r){3-9}
       Dataset & Model & \textbf{Accel.} (\si{m/s^2})& \textbf{Ang. Vel.} (\si{rad/s}) & \textbf{Velocity} (\si{m/s}) & \textbf{Position} (\si{m})& Roll (\si{rad})& Pitch (\si{rad})& Yaw (\si{rad})\\
   \midrule
       \multirow{3}{*}{BeamNGRL} & NoSlip3D  & 19.24 \std{13.97} & 1.89 \std{1.23} & 2.86 \std{2.05} & 22.20 \std{15.09} & 0.24 \std{0.15} & 0.23 \std{0.15} & 1.92 \std{1.93} \\
       & Slip3D  & 7.98 \std{4.71} & 1.19 \std{0.88} & 3.27 \std{2.44} & 11.63 \std{8.80} & 0.20 \std{0.13} & 0.17 \std{0.13} & 1.41 \std{1.63} \\
       & Learned  & 4.79 \std{1.62} & 0.61 \std{0.23} & 2.59 \std{1.10} & 10.39 \std{6.12} & 0.18 \std{0.14} & 0.16 \std{0.12} & 0.96 \std{1.45} \\
    \midrule
       \multirow{3}{*}{Washington} & NoSlip3D  & 7.30 \std{5.48} & 0.59 \std{0.45} & 8.25 \std{7.67} & 28.67 \std{24.82} & 0.18 \std{0.14} & 0.10 \std{0.08} & 0.71 \std{0.96} \\
       & Slip3D  & 5.04 \std{2.06} & 0.34 \std{0.21} & 5.89 \std{3.68} & 16.51 \std{10.89} & 0.18 \std{0.14} & 0.08 \std{0.04} & 0.59 \std{0.76} \\
       & Learned  & 4.57 \std{1.79} & 0.24 \std{0.15} & 2.19 \std{3.24} & 11.36 \std{10.89} & 0.18 \std{0.14} & 0.07 \std{0.05} & 0.40 \std{0.97} \\
    \bottomrule
    \end{tabular}
    \label{tab:model_results}
    \vspace{-10pt}
\end{table*}

We perform a study of three classes of dynamics models over trajectories quantified by our aggressiveness metric \eqref{aggressiveness}. 
The initial dataset $\mathcal{D}$ contains roughly 2000 4-second trajectories in which the vehicle drives over mildly uneven terrain, at average speeds of $8 \pm 1.5~\si{m/s}$ and vertical accelerations of $9.8 \pm 2.7~\si{m/s^2}$. We choose $k_i$ so that the feature vector $\phi$ is conditioned on the entries of the acceleration vector. Each $p_i$ is then fit to $\mathcal{D}$ using a Gaussian defined on the feature space determined by $\phi[i]$. This process allows us to quantify the aggressiveness of additional trajectories relative to $\mathcal{D}$ as outlined in \cref{sec:energy}. The learned model is also trained using $\mathcal{D}$.
The remainder of the dataset's trajectories is collected with speeds between $7-9~\si{m/s}$, traversing shallower and steeper slopes.

Each model $f$ is validated on each of these trajectories (on a held-out set for the learned model) with respect to the ground truth trajectory and the maximum normed error over the horizon ($H$-MNE). Explicitly,
\begin{equation}
    \hmne(f, \tau) = \max_{\vec x^\text{gt}_t\in \tau_{\text{gt}}} \norm{\vec x'_{t} - \vec x_t^{\text{gt}} } \label{hmne}
\end{equation}
where $\vec x'_{t}$ is obtained from rolling out the model from the ground truth initial state, i.e. $\vec x'_{t} = f(\vec x'_{t-1})$ for $t=1,\dots,H$ where $\vec x'_0 \coloneqq \vec x_0^{\text{gt}}$. The maximum is taken over the horizon to reflect the box constraints \eqref{dbox} as \textit{any} violation anywhere is safety-critical.
\begin{figure}[t]
    \vspace{5pt}
    \centering
    \includegraphics[width=\linewidth, trim={0.3cm 0.5cm 0.3cm 0cm}]{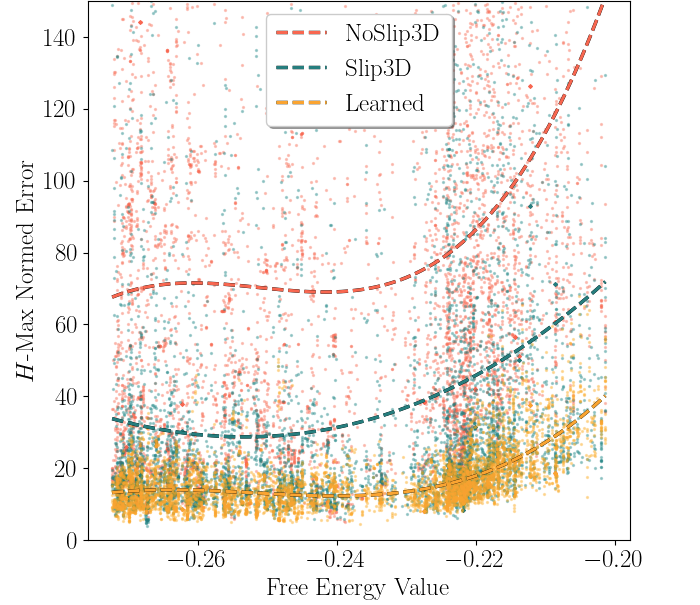}
    \caption{Model performance on validation trajectories. Each marker corresponds to the validation of a model on a single trajectory. A cubic polynomial is fit to each model's performance to show the general trend over increasing Free Energy Values $E(\tau)$. Higher energy indicates more out-of-distribution (more aggressive) relative to the initial dataset.}
    \label{fig:result1}
    \vspace{-15pt}
\end{figure}
The results of this experiment are shown in \cref{fig:result1}. With increasing energy (greater acceleration magnitudes), not only does the performance degrade but the performance \textit{across} models of varying complexity diverge. This empirically confirms our intuition that aggressive driving is a challenging modeling problem with a nontrivial range of unmodeled dynamics.

\subsection{Model Learning with Aggressive Data}

We evaluate the Learned model on two datasets, named the BeamNGRL and Washington datasets. The BeamNGRL dataset is released with this work and its collection process is described in \cref{sec:result1}.

The Washington dataset is collected with a full-size Polaris S4 1000 RZR equipped with IMU, GPS, visual odometry, and a perception network \cite{terrainnet} for terrain mapping. State and sensor messages are synced, or linearly interpolated as necessary, at a rate of 10 Hz to produce the dataset from driving data with speeds up to 14 m/s.
The time horizon for each trajectory is 5 seconds.
The dataset encompasses diverse terrain geometries, ranging from flat surfaces to steep inclines and to banking and twisting trails. Track indents, ditches, and sharp elevation changes also occur alongside these trails. Throughout the dataset, the vehicle additionally interacts with grassy areas, dirt paths, trees, small rocks, logs, bushes and regions densely populated with vegetation.

The timestep size for each trajectory in both datasets is 0.1 seconds.
The results of the models are shown in \Cref{tab:model_results}. The numbers displayed are $H$-MNE errors computed through \eqref{hmne}. Vector states (i.e. acceleration, angular velocity, velocity, position) are demarcated in bold text.
It can be observed in table \ref{tab:model_results} that with increasing model complexity, the errors go down.
It can also be observed that there exists a large gap in position error between the NoSlip3D and Slip3D models for the BeamNGRL dataset, implying that accounting for wheel slippage addresses most of this error.
The velocity errors in the BeamNGRL dataset are higher for Slip3D than for NoSlip3D due to the first-order integration of small errors in acceleration.
The learned model does not suffer the same fate as Slip3D in velocity predictions due to lower acceleration and angular velocity errors.
Note that the errors in velocity for the real dataset may be higher due to errors in system identification, as we assume vehicle parameters such as friction to be constant throughout the dataset.
It is also observed that the gap between the models is much smaller for the Washington dataset than it is for the synthetic dataset from BeamNG \cite{beamng_tech}.
This is a consequence of the driving in the synthetic dataset being far more aggressive, as there are no safety concerns for the driver in the simulation. We provide video demonstrations to show examples of simulated low energy \footnote{\begin{small}\url{https://tinyurl.com/knzy8kzn}\end{small}}, medium energy\footnote{\begin{small}\url{https://tinyurl.com/2p993ktu}\end{small}}, and high energy\footnote{\begin{small}\url{https://tinyurl.com/ynx8rbhx}\end{small}} driving.

\section{Conclusion}

Using BeamNG, we collect a realistic and diverse aggressive off-road driving dataset. This dataset is used in a quantitative study on the effect of aggressiveness on models with various modeled dynamics. An energy score conditioned on higher-order states is proposed for quantifying the aggressiveness of trajectories. Our results show that model predictions degrade with increasing aggressiveness and that simpler models degrade faster.

Next, we evaluate the performance of these dynamics models on two different off-road datasets; the BeamNGRL and the Washington dataset. Importantly, the models are validated on their accuracy over higher-order states, which are critical in the aggressive regime.

We acknowledge that the definitions and metrics provided in this work can be further developed. For instance, semantic or perceptual information (e.g. grass, dirt, gravel) is readily available through state-of-the-art perception networks \cite{terrainnet, frey_roadrunner_2024}. This information can better differentiate operating regimes and improve the performance of dynamics models. We aim to incorporate this in future work.

\clearpage
\newpage

\printbibliography

\end{document}